\documentclass[conference]{IEEEtran}
\usepackage{cite}
\usepackage{graphicx}
\usepackage{amsmath,amssymb,amsfonts}
\usepackage[table]{xcolor}
\usepackage{algorithm}
\usepackage{balance}
\usepackage{algpseudocode}
\usepackage[inline]{enumitem}
\usepackage{placeins}
\usepackage{booktabs}
\usepackage{multirow}
\usepackage{subcaption}
\usepackage{tikz}
\usetikzlibrary{arrows.meta,positioning,fit,shapes.misc,calc}
\usepackage{subcaption}
\usepackage[font=small,skip=4pt]{caption}
\usepackage{breakurl}

\title{Machine Learning for LiDAR-Based Indoor Surface Classification in Intelligent Wireless Environments}

\author{
    \IEEEauthorblockN{
        Parth Ashokbhai Shiroya,
        Swarnagowri Shashidhar,
        Amod Ashtekar, 
        Krishna Aindrila Kar, \\ Rafaela Lomboy, Dalton Davis$^*$,  and 
        Mohammed E. Eltayeb}
    \IEEEauthorblockA{
        Department of Electrical and Electronics Engineering \\
        California State University, Sacramento,  Sacramento,  CA, USA \\
        $^*$Department of Computer Science\\
        University of California, Davis, Davis, CA, USA \\
     Email:parthshiroya@csus.edu
    }
}

\begin{document}
\maketitle

\begin{abstract}
Reliable connectivity in millimeter-wave (mmWave) and sub-terahertz (sub-THz) networks depends on reflections from surrounding surfaces, as high-frequency signals are highly vulnerable to blockage. The scattering behavior of a surface is determined not only by material permittivity but also by roughness, which governs whether energy remains in the specular direction or is diffusely scattered. This paper presents a LiDAR-driven machine learning framework for classifying indoor surfaces into semi-specular and low-specular categories, using optical reflectivity as a proxy for electromagnetic scattering behavior. A dataset of over 78,000 points from 15 representative indoor materials was collected and partitioned into $3~\text{cm} \times 3~\text{cm}$ patches to enable classification from partial views. Patch-level features capturing geometry and intensity, including elevation angle, natural-log–scaled intensity, and max-to-mean ratio, were extracted and used to train Random Forest, XGBoost, and neural network classifiers. Results show that ensemble tree-based models consistently provide the best trade-off between accuracy and robustness, confirming that LiDAR-derived features capture roughness-induced scattering effects. The proposed framework enables the generation of scatter-aware environment maps and digital twins, supporting adaptive beam management, blockage recovery, and environment-aware connectivity in beyond-5G and 6G networks.
\end{abstract}

\begin{IEEEkeywords}
Machine learning, LiDAR sensing, Surface roughness classification, mmWave,  tera-Hertz, Smart environments.
\end{IEEEkeywords}

\section{Introduction}

Future wireless systems are moving to millimeter-wave (mmWave) and sub-terahertz (sub-THz) bands to support ultra-high data rates and low-latency connectivity \cite{Rappaport2019B5G}. At these frequencies, signals are highly susceptible to blockage, and non-line-of-sight (NLoS) connectivity depends on reflections from surrounding surfaces, including natural objects and engineered structures such as reconfigurable intelligent surfaces (RIS) and passive reflectors \cite{SpaceBeam,Ewe2018Density,giordani2020toward6g}. Accurate knowledge of environmental scattering behavior, particularly whether surfaces act as specular reflectors or diffuse scatterers, is therefore essential for link planning, blockage mitigation, and adaptive beam management in mmWave and sub-THz networks.

Recent 6G roadmaps identify integrated sensing and communications (ISAC) and intelligent wireless environments (IWE) as key enablers for addressing these challenges \cite{Gonzalez2024ISAC,khanIRSsurvey2025}. Measurement studies confirm that reflection strength depends strongly on material composition, roughness, and geometry \cite{SubTHzMeasVTC2024,cui2024terahertz,ju2019scattering,Rappaport2015,Langen1994}. In parallel, multimodal sensing frameworks have shown the benefits of out-of-band modalities such as LiDAR, cameras, and radar for reducing beam training overhead \cite{MILCOM2024Multimodal,VTC2024Multimodal}. However, prior work has largely focused on beam tracking or environment reconstruction, while the fundamental problem of determining the scattering properties of a surface or an environment remains underexplored.  

LiDAR offers a promising approach to bridge this gap. Beyond capturing 3D geometry, LiDAR intensity encodes optical reflectivity, which is influenced by both material properties and surface roughness, and can serve as a proxy for scattering behavior at high frequencies. Prior studies have applied LiDAR to localization \cite{Karimi2022LoLaSLAM}, multimodal beam prediction \cite{Klautau2018LidarDLBeamSelection,davis2025lidar}, and material classification with multispectral or polarimetric extensions \cite{Han2021PolarimetricLiDAR,Farhani2021LidarClassification}. More recent work has shown that LiDAR can even predict mmWave beam directions \cite{SpaceBeam}. Yet most existing methods assume dense scans or detailed 3D models, which are impractical in real-time deployments.  

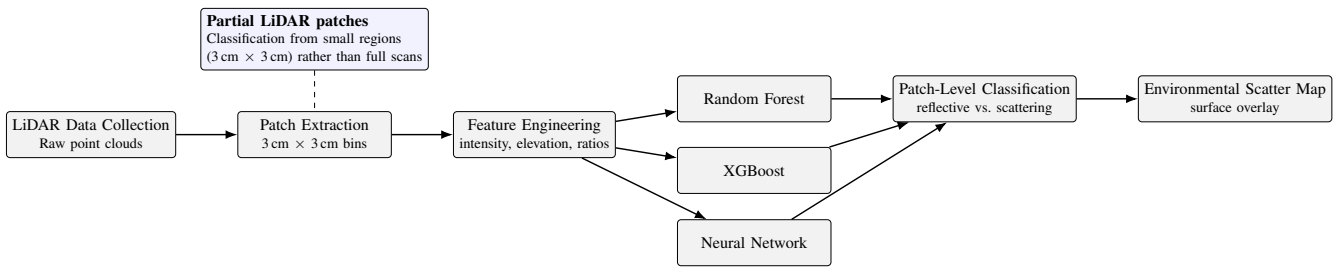
\begin{figure*}[t]
\centering
\resizebox{2\columnwidth}{!}{
\begin{tikzpicture}[
  font=\small,
  node distance=10mm and 12mm,
  >=Latex,
  block/.style={draw, rounded corners=2pt, fill=gray!10, align=center, minimum width=30mm, minimum height=9mm},
  line/.style={-Latex, thick},
  callout/.style={draw, rounded corners=2pt, fill=blue!5, align=left}
]

\node[block] (lidar) {LiDAR Data Collection\\ \footnotesize Raw point clouds};
\node[block, right=of lidar] (patch) {Patch Extraction\\ \footnotesize 3\,cm $\times$ 3\,cm bins};
\node[block, right=of patch] (features) {Feature Engineering\\ \footnotesize intensity, elevation, ratios};
\node[block, right=of features, yshift=7mm] (rf) {Random Forest};
\node[block, right=of features, yshift=-7mm] (xgb) {XGBoost};
\node[block, below=19mm of rf] (nn) {Neural Network};
\node[block, right=12mm of rf] (class) {Patch-Level Classification\\ \footnotesize reflective vs.\ scattering};
\node[block, right=of class] (scatter) {Environmental Scatter Map\\ \footnotesize surface overlay};

\draw[line] (lidar) -- (patch);
\draw[line] (patch) -- (features);
\draw[line] (features) -- (rf);
\draw[line] (features) -- (xgb);
\draw[line] (features) -- (nn);
\draw[line] (rf) -- (class);
\draw[line] (xgb) -- (class);
\draw[line] (nn) -- (class);
\draw[line] (class) -- (scatter);

\node[callout, above=8mm of patch, align=left] (call)
{\textbf{Partial LiDAR patches}\\
\footnotesize Classification from small regions\\
\footnotesize (3\,cm $\times$ 3\,cm) rather than full scans};
\draw[dashed] (call.south) -- (patch.north);

\end{tikzpicture}}
\caption{Proposed workflow: LiDAR point clouds are partitioned into small patches (3\,cm $\times$ 3\,cm). Patch-level features are fed into three complementary ML models, namely,  Random Forest (robust to small datasets), XGBoost (strong gradient-boosting baseline), and a lightweight Neural Network (capturing nonlinear feature interactions), to classify each patch as semi-specular (reflective) or low-specular (scattering). Multiple models are evaluated to benchmark performance and identify the most effective approach. In future deployment, the best-performing model can be used to aggregate patch-level predictions into environmental scatter maps for intelligent wireless environments.}
\label{fig:block-diagram}
\end{figure*}

\textbf{This paper addresses this gap.} We propose a LiDAR-driven machine learning framework that captures surface-dependent scattering signatures and classifies indoor surfaces as reflective (semi-specular) or scattering (low-specular) using only small local patches ($3~\text{cm}\times3~\text{cm}$). Using scans of 15 representative materials, we extract geometric and intensity descriptors and evaluate Random Forest, XGBoost, and a lightweight neural network under a leave-surface-out protocol. We identify machine-learnable descriptors that enable LiDAR-based surface classification and show that tree-based ensembles provide the best balance of accuracy and robustness. These results lay the groundwork for constructing scatter-aware environment maps and digital twins from LiDAR alone, which can support adaptive beam management and environment-aware communication in next-generation mmWave and sub-THz systems.

\section{Motivation and Framework}

The objective of this work is to develop a sensing-driven framework that leverages LiDAR-derived features to classify indoor surfaces into reflective and scattering categories. Such classification is essential for environment-aware modeling in mmWave and sub-THz systems, where non-line-of-sight connectivity depends strongly on reflection strength. Surface behavior at these frequencies is not determined by material permittivity alone, but also by micro-scale roughness. According to the Rayleigh criterion, a surface is considered rough if its root-mean-square height $h_{\text{rms}}$ exceeds the critical threshold $h_c = \frac{\lambda}{8\cos\theta_i}$, where $\lambda$ is the wavelength and $\theta_i$ is the angle of incidence \cite{ju2019scattering}. When this condition is met, specular reflection is significantly reduced and energy is scattered into diffuse directions.  The loss of coherent reflection due to roughness can be modeled by the scattering factor $
\rho_s = \exp\left[-8 \left(\frac{\pi h_{\text{rms}} \cos\theta_i}{\lambda}\right)^2\right],$
which directly attenuates the Fresnel reflection coefficients.  This relationship highlights that even highly reflective materials (e.g., metal) may yield weak or diffuse returns if their microtexture introduces strong scattering. For wireless systems, this implies that surface-dependent scattering properties play a decisive role in determining whether a wall, ceiling, or furnishing contributes to useful reflections or degrades link quality.  

LiDAR sensing offers a unique opportunity to capture these surface-dependent effects. By acquiring both geometric structure and return intensity, LiDAR encodes indirect information about roughness-induced scattering. Our proposed framework, illustrated in Fig.~\ref{fig:block-diagram}, processes LiDAR scans of 15 representative indoor materials. Each scan is partitioned into $3~\mathrm{cm} \times 3~\mathrm{cm}$ patches, enabling classification from partial views rather than requiring exhaustive coverage. For each patch, we extract discriminative descriptors that capture both geometric and reflectivity properties, including mean elevation angle, natural-log–scaled intensity \((\ln(1+x))\), and the max-to-mean intensity ratio. These features are then provided as input to supervised learning models,  Random Forest, XGBoost, and a lightweight neural network, to assess which learning paradigm is most effective for LiDAR-based reflectivity classification.  Our central hypothesis is that geometric and intensity features extracted from small LiDAR patches are sufficient to discriminate reflective from scattering surfaces, as they implicitly capture roughness-induced losses predicted by electromagnetic theory.

\section{Experimental Setup and Data Acquisition}
\begin{figure}[t]
\centering
\vspace{-10mm}
\includegraphics[width=\linewidth]{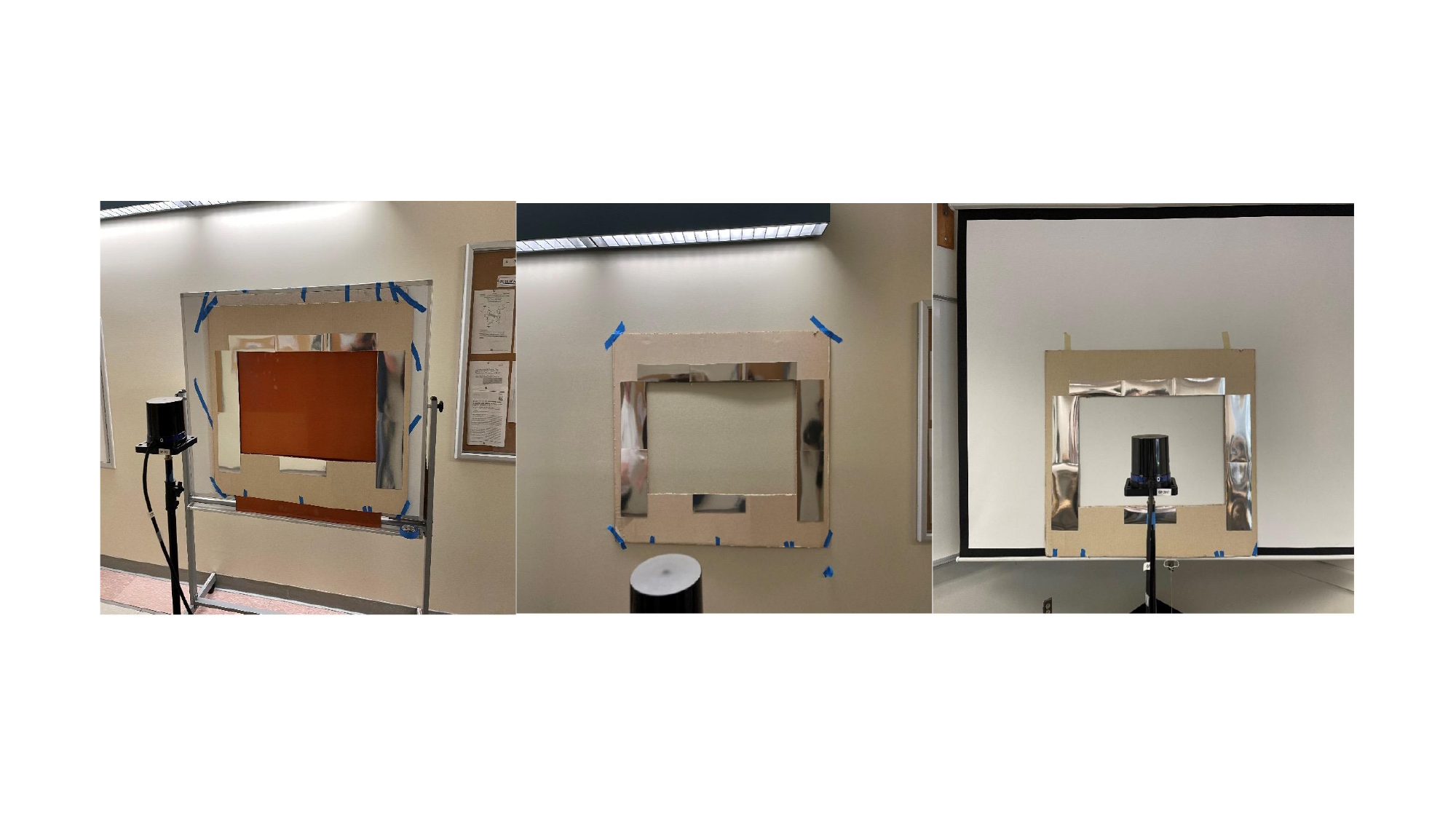}
\vspace{-15mm}
\caption{Experimental setup showing copper, drywall, and projector screen surfaces mounted on a cardboard frame. The Quanergy M8 LiDAR sensor captured point clouds used for patch-based reflectivity analysis.}
\label{fig:combined_surfaces}
\end{figure}

The dataset was acquired using a Quanergy M8 LiDAR sensor equipped with eight lasers that generate eight scanning rings. To ensure consistency, each surface under test was mounted on a cardboard frame with reflective boundaries ($34 \times 18$ inches) to define a uniform region of interest (ROI). The sensor, placed on a tripod 44 inches away, was vertically aligned such that the bottom scanning ring coincided with the frame edge. For each material, a 10-second scan was recorded and one representative frame was extracted. An example experimental setup with three representative materials (copper, drywall, and projector screen) is shown in Fig.~\ref{fig:combined_surfaces}.  The raw point clouds were cropped in CloudCompare to isolate the ROI and exported to CSV format. Each ROI was further partitioned into $3~\mathrm{cm} \times 3~\mathrm{cm}$ patches, yielding 1,339 patches across 15 distinct indoor materials. This patch-level representation enables classification from partial scans rather than requiring exhaustive coverage of entire surfaces.

LiDAR intensity values were linearized to compute a specularity metric defined as $\mathrm{Specularity~(dB)} = 10 \log_{10}
\left(
\frac{\max(x_n)}{\mathrm{mean}(x_n)}
\right),$ where $x_n$ denotes the linearized intensity values of points in a patch. The metric captures the ratio between peak and average return intensity, thereby emphasizing sharp reflections. Higher specularity values indicate smooth, highly reflective surfaces, while lower values correspond to rough or scattering surfaces. This metric has also been adopted in prior work such as \cite{SpaceBeam}.

Table~\ref{tab:sheet_point_patch_counts} summarizes the dataset composition. In total, 78,165 LiDAR points were collected and partitioned into 1,339 patches. The dataset spans diverse indoor materials and was segregated into two classes (semi-specular and low-specular) using the specularity metric with a threshold of $\mathrm{Specularity~(dB)} >10\,\mathrm{dB}$.  To support reproducibility and further research, this dataset has been made publicly available at GitHub repository: \texttt{github.com/AmodAshtekar/surface-data}

\section{Learning-Based Surface Classification}
\label{sec:ml_framework}

Machine learning provides a natural framework for classifying surface scattering behavior from LiDAR point cloud data, which includes spatial coordinates, distance, and intensity (or power) returns. Different materials interact with LiDAR in distinct ways. Smooth surfaces often generate sharp specular-like intensity peaks, while rough or absorptive surfaces produce more diffuse scattering patterns. By extracting geometric and intensity-based descriptors from local patches, a classifier can learn to separate these scattering signatures and determine whether a surface behaves in a semi-specular or low-specular manner. 

In this work, we focus on \emph{semi-specular} surfaces, in contrast to highly specular ones such as mirrors and strongly diffuse surfaces. For highly specular materials, the LiDAR return often encodes secondary reflections rather than intrinsic material response, which makes the signature unreliable. For example, a mirror patch may primarily reflect the profile of a neighboring wall. Because of this limitation, our analysis concentrates on distinguishing semi-specular from low-specular materials, where LiDAR measurements more reliably capture material-dependent scattering.

\subsection{Dataset and Patch Representation}

\subsubsection{Surface Labeling and Patch Formation}
Each surface was pre-labeled with its ground-truth class. To mitigate noise and local variability, point-level measurements were aggregated into $3\,\mathrm{cm} \times 3\,\mathrm{cm}$ patches. This resolution was chosen after testing multiple bin sizes. Smaller patches below $2$ cm produced noisy results due to jitter, while larger ones above $5$ cm suppressed surface-specific variation.  Table~\ref{tab:sheet_point_patch_counts} summarizes the number of points and patches across all surfaces.

\subsubsection{Feature Construction}
We initially explored a broad set of patch-level descriptors, including raw attributes (e.g., mean linearized intensity, elevation angle, log-scaled intensity), derived features (e.g., max-to-mean ratio, ratio of log-intensity to incidence angle), and neighborhood statistics across LiDAR scanning rings.  

After evaluation, we found that a compact subset of three features consistently provided the best discriminative power while avoiding redundancy.  
The first is the mean elevation angle, which captures geometric orientation.  
The second is the maximum log-scaled intensity within a patch, defined as
\[
\texttt{log\_linear\_max} = \max_{p \in \text{patch}} (\log\_linear_p),
\]
where $\log\_linear_p = \ln(1 + \texttt{linear}_p)$ is the log-scaled intensity of point $p$.

In practice, NumPy’s \texttt{np.log1p} function was used for numerical stability and to safely handle zeros.  
We selected the natural logarithm rather than base-10 or base-2 logs because it is standard in machine learning preprocessing; the base choice only introduces a constant scaling factor and does not affect classification outcomes.  
The third feature is the max-to-mean ratio (MMR), defined as
\[
\text{MMR} = \frac{\texttt{log\_linear\_max}}{\texttt{log\_linear\_mean} + \varepsilon},
\]
where $\texttt{log\_linear\_max}$ is the maximum log-scaled intensity within a patch and
$\texttt{log\_linear\_mean}$ is the mean of the log-scaled intensities across all points in the patch.

Together, these three descriptors — mean elevation angle, log-linear-max (log-scaled maximum intensity), and MMR — formed a compact and complementary feature set that achieved the strongest performance in distinguishing semi-specular from low-specular surfaces.

\begin{table}[t]
\renewcommand{\arraystretch}{1.2}
\centering
\caption{LiDAR point and patch counts per surface. Semi-specular materials (Specularity metric \(>10\,\mathrm{dB}\)) are metal copper, metal tin, whiteboard, linoleum, and TV screen; the remaining materials are labeled low-specular.}
\label{tab:sheet_point_patch_counts}
\begin{tabular}{lrr}
\toprule
\textbf{Material} & \textbf{Total Points} & \textbf{Patches} \\
\midrule
Metal copper     & 4,896 & 143 \\
Metal tin        & 4,780 & 94  \\
Whiteboard       & 4,940 & 79  \\
Projector screen & 4,994 & 75  \\
TV               & 4,993 & 85  \\
Linoleum         & 5,189 & 82  \\
Smooth wood      & 5,073 & 80  \\
Rough wood       & 5,048 & 84  \\
Drywall          & 5,032 & 77  \\
Cardboard        & 5,010 & 80  \\
Corkboard        & 4,965 & 81  \\
Styrofoam        & 4,977 & 72  \\
Concrete wall    & 5,025 & 83  \\
Fabric pinboard  & 4,984 & 78  \\
Carpet           & 3,193 & 96  \\
\midrule
\textbf{Total}   & \textbf{78,165} & \textbf{1,339} \\
\bottomrule
\end{tabular}
\end{table}

\begin{figure}[t]
\centering
\begin{subfigure}[b]{0.50\linewidth}
  \includegraphics[width=\linewidth]{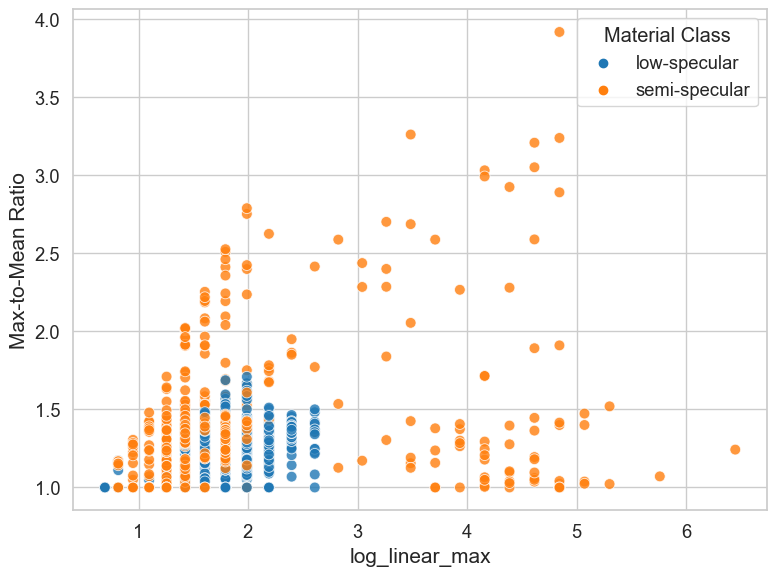}
  \caption{\small \texttt{log\_linear\_max} vs MMR}
  \label{fig:features_combo1}
\end{subfigure}\hfill
\begin{subfigure}[b]{0.50\linewidth}
  \includegraphics[width=\linewidth]{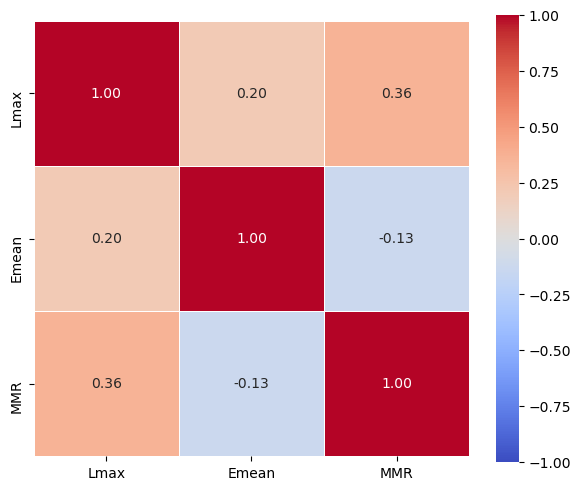}
  \caption{\small Correlation}
  \label{fig:features_combo2}
\end{subfigure}
\caption{Feature behavior and complementarity. Semi-specular patches skew to higher peaks and MMR,  and  moderate to low correlations support joint use in classification models.}
\label{fig:features_combo}
\end{figure}

As shown in Fig.~\ref{fig:features_combo1}, semi-specular patches cluster at higher maximum intensity and max-to-mean ratio values, while low-specular patches remain broadly distributed with lower values. This confirms that intensity peaks and their relative strength are key discriminative features.   The heatmap in Fig.~\ref{fig:features_combo2}, confirms that the key features are only moderately correlated, which indicates that they provide complementary rather than redundant information. Taken together, the clustering behavior and correlation structure demonstrate that the selected feature subset forms a compact and discriminative representation of surface scattering properties.

%



\subsection{Classification Framework}

\subsubsection{Training and Evaluation Protocol} \label{sec:training_strategy}
To enforce generalization, training and testing were conducted on disjoint sets of material surfaces.  Four surfaces, two semi-specular and two low-specular, were fixed as the test set. For a given training size $k$, we randomly sampled $k$ additional surfaces (materials) to form the training set, using all patches from those surfaces. Testing was always performed on the fixed unseen surfaces. This leave-surface-out protocol was repeated 50 times for each $k$, ensuring statistical robustness. Normalization was applied only to the Neural Network using a \texttt{StandardScaler}, while Random Forest and XGBoost operated on raw features.

\subsubsection{Model Architectures}
Three supervised learning models were implemented: Random Forest, XGBoost~\cite{Chen2016XGBoost}, and a feed-forward Neural Network. Hyperparameters for the ensemble models were tuned by grid search following established practices~\cite{Probst2019RF}, with final values listed in Table~\ref{tab:hyperparams}.  
The Neural Network consisted of two hidden layers with 64 and 32 ReLU units, dropout regularization with rate 0.3, and a softmax output layer for binary classification. Training used the Adam optimizer with categorical cross-entropy loss, mini-batches of 32, and early stopping on a validation set to prevent overfitting~\cite{Zhang2019MobileDL}.

\subsubsection{Evaluation Metrics}
Performance was assessed using accuracy, precision, recall, F1-score, and confusion matrices. Because the dataset was moderately imbalanced, we also used precision–recall curves, which provide more reliable insight than ROC curves under such conditions~\cite{Saito2015PR}. All evaluations were performed on held-out test surfaces, ensuring that the reported results reflect generalization to unseen materials.

\subsection{Model Performance and Comparison}

\begin{figure}[t]
    \centering
    \includegraphics[width=0.48\textwidth]{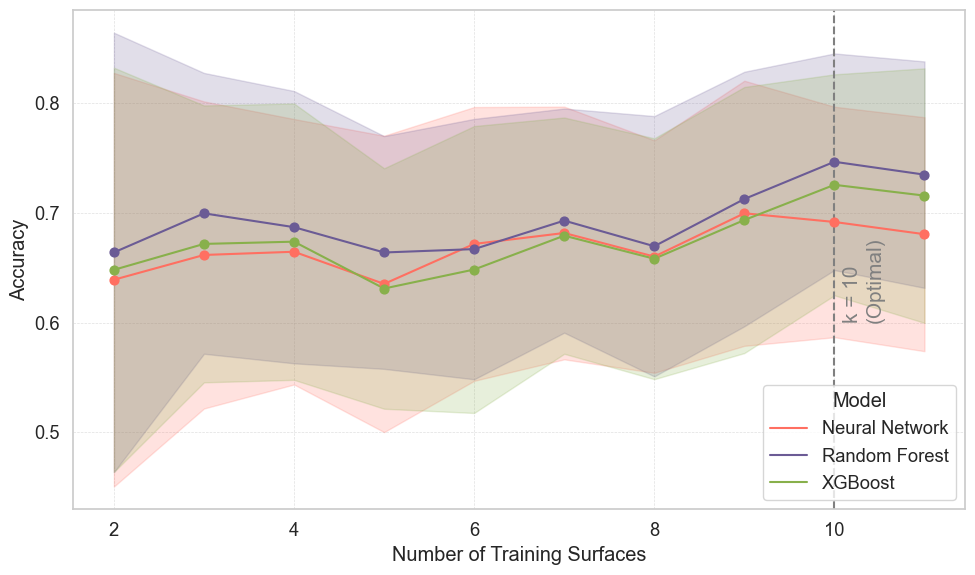}
    \caption{Classification accuracy vs. number of training surfaces. Random Forest and XGBoost outperform Neural Network across most settings. Error bars show standard deviation over 50 runs.}
    \label{fig:AccuracyVsTrainingSheets}
\end{figure}

Across training sizes $k$, Random Forest and XGBoost consistently outperformed the Neural Network (see Fig.~\ref{fig:AccuracyVsTrainingSheets}). Random Forest delivered the highest mean accuracy with the lowest variance, while XGBoost occasionally reached higher peaks but with greater variability. The Neural Network was competitive for small $k$ but lacked robustness as $k$ increased. Overall, Random Forest offered the best balance between accuracy and stability, with XGBoost a close second. Table~\ref{tab:acc_all} summarizes model accuracies across training sizes.

\section{Experimental Results and Analysis}
\label{sec:results}

All experiments followed the leave-surface-out protocol described in Section~\ref{sec:training_strategy}, which ensured that training and testing surfaces were always disjoint. Results are reported at the patch level using accuracy, precision, recall, and F1-score as evaluation metrics, with confusion matrices generated to visualize class-specific errors. Patch-level evaluation is particularly relevant for LiDAR-based classification because it reflects realistic scanning conditions in which only fragments of a surface are observed due to occlusion, angle restrictions, or incomplete coverage. Each $3\,\mathrm{cm} \times 3\,\mathrm{cm}$ patch is classified independently, forcing the model to recognize local scattering characteristics rather than memorizing entire surfaces.

\begin{table}[t]
\centering
\caption{Final hyperparameters for XGBoost and Random Forest.}
\label{tab:hyperparams}
\begin{tabular}{lll}
\toprule
\textbf{Model} & \textbf{Parameter} & \textbf{Value} \\
\midrule
XGBoost & n\_estimators & 100 \\
        & max\_depth & 5 \\
        & learning\_rate & 0.1 \\
        & subsample & 0.8 \\
        & colsample\_bytree & 0.8 \\
        & reg\_alpha, reg\_lambda & 0.1, 1 \\
        & gamma & 1 \\
\midrule
Random Forest & n\_estimators & 200 \\
              & max\_depth & 10 \\
              & max\_features & sqrt \\
              & min\_samples\_split & 5 \\
              & min\_samples\_leaf & 2 \\
              & class\_weight & balanced \\
\bottomrule
\end{tabular}
\end{table}

\begin{table*}[t]
\centering
\caption{Summary of maximum, mean, and standard deviation of accuracy across models for different training sizes ($k$). Results are averaged over 50 runs. Bold indicates best values.}
\label{tab:acc_all}
\setlength{\tabcolsep}{4pt}
\footnotesize
\begin{tabular*}{\textwidth}{@{\extracolsep{\fill}} c ccc ccc ccc }
\toprule
& \multicolumn{3}{c}{\textbf{Maximum Accuracy}} & \multicolumn{3}{c}{\textbf{Mean Accuracy}} & \multicolumn{3}{c}{\textbf{Std. Dev.}} \\
\cmidrule(lr){2-4} \cmidrule(lr){5-7} \cmidrule(lr){8-10}
\textbf{$k$} & NN & RF & XGB & NN & RF & XGB & NN & RF & XGB \\
\midrule
2  & \textbf{0.974} & 0.967 & 0.956 & 0.639 & \textbf{0.664} & 0.648 & 0.188 & 0.200 & \textbf{0.184} \\
3  & 0.942 & \textbf{0.960} & 0.957 & 0.662 & \textbf{0.699} & 0.672 & 0.140 & 0.128 & \textbf{0.126} \\
4  & \textbf{0.935} & 0.948 & 0.930 & 0.664 & \textbf{0.687} & 0.674 & \textbf{0.121} & 0.124 & 0.126 \\
5  & \textbf{0.951} & 0.861 & 0.905 & 0.635 & \textbf{0.664} & 0.631 & 0.135 & \textbf{0.106} & 0.110 \\
6  & 0.932 & 0.908 & \textbf{0.914} & \textbf{0.672} & 0.667 & 0.648 & \textbf{0.125} & 0.119 & 0.131 \\
7  & 0.906 & \textbf{0.933} & \textbf{0.933} & 0.682 & \textbf{0.693} & 0.679 & 0.115 & \textbf{0.102} & 0.108 \\
8  & \textbf{0.941} & 0.904 & 0.897 & 0.660 & \textbf{0.669} & 0.658 & \textbf{0.106} & 0.119 & 0.110 \\
9  & \textbf{0.982} & 0.963 & 0.960 & 0.700 & \textbf{0.713} & 0.693 & \textbf{0.121} & \textbf{0.116} & \textbf{0.121} \\
10 & 0.859 & \textbf{0.910} & 0.907 & 0.692 & \textbf{0.746} & 0.725 & 0.105 & \textbf{0.099} & 0.101 \\
11 & 0.832 & \textbf{0.914} & 0.889 & 0.680 & \textbf{0.735} & 0.716 & \textbf{0.107} & \textbf{0.103} & 0.116 \\
\bottomrule
\end{tabular*}
\end{table*}

\begin{figure}[t]
    \centering
    \includegraphics[width=0.48\textwidth]{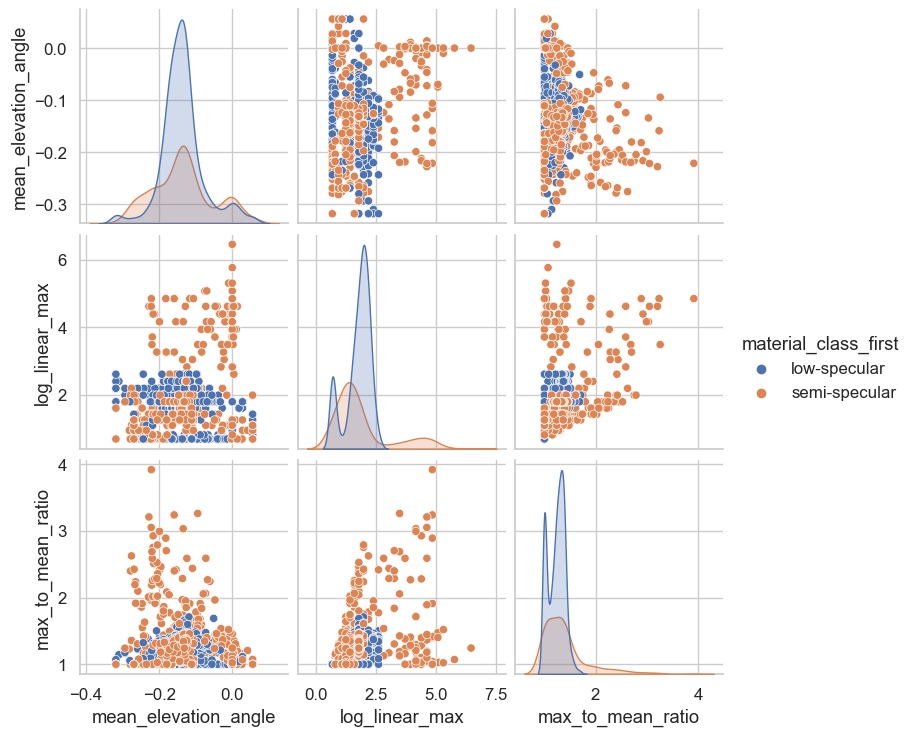}
    \caption{Pairwise distributions of the three final features (\texttt{mean\_elevation\_angle}, \texttt{log\_linear\_max} (natural-log scale, \(\ln(1+x)\)),  \texttt{max\_to\_mean\_ratio}) by class.  Semi-specular patches exhibit shifts to higher intensity-related values, while \texttt{mean\_elevation\_angle} contributes complementary orientation information.}
    \label{fig:pairplot}
\end{figure}

To assess the separability of semi-specular and low-specular surfaces, we first examined the distributions of the three most discriminative features: \texttt{mean\_elevation\_angle}, \texttt{log\_linear\_max}, and \texttt{max\_to\_mean\_ratio}. As shown in Fig.~\ref{fig:pairplot}, semi-specular patches consistently shift toward higher \texttt{log\_linear\_max} and \texttt{max\_to\_mean\_ratio} values, while low-specular patches remain broadly distributed at lower intensities. Off-diagonal scatterplots reveal that the combination of \texttt{log\_linear\_max} and \texttt{max\_to\_mean\_ratio} provides the clearest class separation, whereas \texttt{mean\_elevation\_angle} is less discriminative in isolation but provides complementary orientation cues when combined with intensity-based features. These observations align with the feature importance ranking obtained from Random Forest and explain why the models are able to distinguish the two surface categories.

\begin{figure}[t]
\centering
\begin{subfigure}[b]{0.49\linewidth}
  \includegraphics[width=\linewidth]{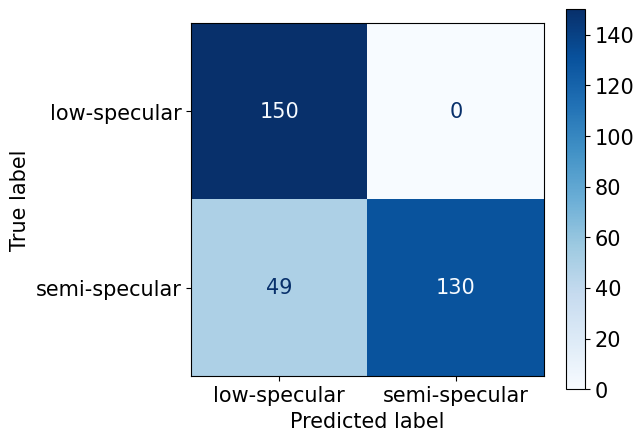}
  \caption{\footnotesize RF}
\end{subfigure}\hfill
\begin{subfigure}[b]{0.49\linewidth}
  \includegraphics[width=\linewidth]{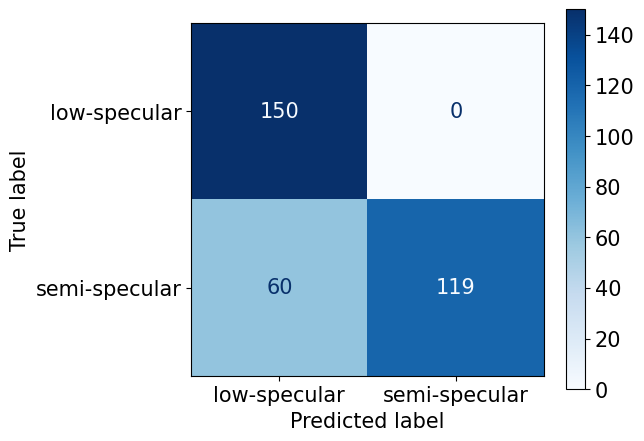}
  \caption{\footnotesize XGB}
\end{subfigure}

\vspace{2pt}
\begin{subfigure}[b]{0.49\linewidth}
  \includegraphics[width=\linewidth]{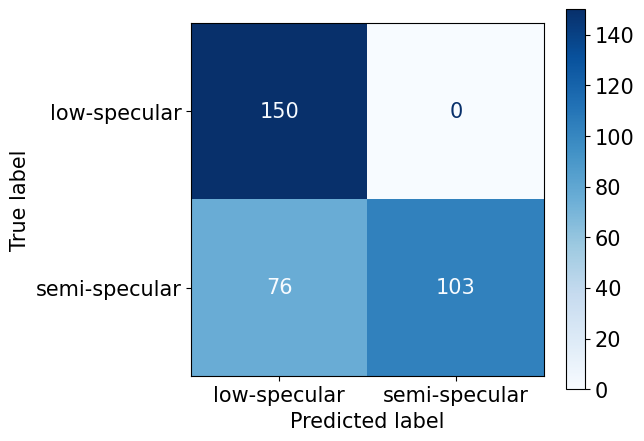}
  \caption{\footnotesize NN}
\end{subfigure}
\vspace{-2mm}
\caption{Patch-level confusion matrices on unseen surfaces.}
\label{fig:conf_mats}
\vspace{-3mm}
\end{figure}

To further investigate class-specific performance, confusion matrices were generated for the three classifiers, shown in Fig.~\ref{fig:conf_mats}. Training was performed on ten surfaces, including \texttt{smooth\_wood}, \texttt{concrete\_wall}, \texttt{rough\_wood}, \texttt{cardboard}, \texttt{drywall}, \texttt{corkboard}, \texttt{projector\_screen}, \texttt{linoleum}, \texttt{carpet}, and \texttt{metal\_copper}. Testing was conducted on four completely unseen surfaces: \texttt{metal\_tin}, \texttt{tv}, \texttt{styrofoam}, and \texttt{fabric\_pinboard}. Each confusion matrix reflects patch-level classification outcomes, with one decision made per $3\,\mathrm{cm} \times 3\,\mathrm{cm}$ patch.  
The results show that all three models achieve perfect classification of low-specular patches, correctly identifying all $150/150$ test samples. Their performance diverges, however, for semi-specular patches. Random Forest performs best, correctly classifying 130 out of 179, followed by XGBoost with 119 and Neural Network with 103. This gap reflects the inherent ambiguity of semi-specular surfaces, whose intermediate scattering behavior makes them more challenging to separate from diffuse materials.

%

The per-class precision, recall, and F1-scores are summarized in Table~\ref{tab:classification_report}. Random Forest achieves the best balance, with the highest overall accuracy (0.84) and strong recall on semi-specular patches, while also maintaining high precision for low-specular patches. XGBoost performs similarly but with slightly weaker recall on semi-specular data. The Neural Network achieves lower accuracy (0.77) and struggles most with semi-specular classification, indicating limited generalization despite competitive peak accuracy at small training sizes.

In summary,  results demonstrate that intensity-related features provide strong separability between semi-specular and low-specular surfaces, and that Random Forest offers the most reliable balance between accuracy, stability, and per-class generalization. XGBoost is a competitive alternative, while the Neural Network shows promise at small training sizes but lacks robustness. These findings suggest that tree-based ensemble methods are best suited for practical LiDAR-based material classification, particularly when robustness to noise and incomplete data is required.

\begin{table}[t]
\centering
\caption{Classification metrics for each model (patch-level evaluation).}
\label{tab:classification_report}
\begin{tabular}{llccc}
\toprule
\textbf{Model} & \textbf{Class} & \textbf{Precision} & \textbf{Recall} & \textbf{F1-score} \\
\midrule
\multirow{2}{*}{XGBoost}
& Low-specular & 0.71 & 1.00 & 0.83 \\
& Semi-specular & 1.00 & 0.66 & 0.80 \\
\cmidrule{2-5}
& \textbf{Accuracy} & \multicolumn{3}{c}{0.82} \\
\midrule
\multirow{2}{*}{Random Forest}
& Low-specular & 0.75 & 1.00 & 0.86 \\
& Semi-specular & 1.00 & 0.73 & 0.84 \\
\cmidrule{2-5}
& \textbf{Accuracy} & \multicolumn{3}{c}{0.84} \\
\midrule
\multirow{2}{*}{Neural Network}
& Low-specular & 0.66 & 1.00 & 0.80 \\
& Semi-specular & 1.00 & 0.58 & 0.73 \\
\cmidrule{2-5}
& \textbf{Accuracy} & \multicolumn{3}{c}{0.77} \\
\bottomrule
\end{tabular}
\end{table}

\section{Conclusion and Future Work}

This paper introduced a LiDAR-based machine learning framework for classifying indoor surfaces into semi-specular and low-specular categories. Such classification is critical for planning and optimizing high-frequency wireless communication links, particularly in mmWave and THz systems where scattering strongly influences coverage and reliability.  A controlled indoor campaign with a Quanergy M8 LiDAR sensor produced a structured dataset of more than 78,000 points across 15 representative materials. Using compact patch-level features extracted from LiDAR point clouds, we showed that surface scattering behavior can be reliably inferred and used to guide link planning.  Among the evaluated models, Random Forest consistently achieved the best balance of accuracy and robustness, underscoring the suitability of ensemble methods for material recognition in realistic propagation environments.   Although the dataset was limited to 15 surfaces and a single LiDAR sensor, the results highlight the potential of LiDAR-derived features for constructing radio environment maps and digital twins that enable adaptive beam management and blockage recovery. Future work will expand the framework to multi-class material classification, incorporate complementary sensing modalities such as radar, and implement real-time operation in smart infrastructure platforms with the objective of advancing environment-aware communication strategies for 6G and beyond  networks.

\section*{Acknowledgment}
This material is based on work supported by the National Science Foundation under grant No. NSF-2243089.

\balance

\bibliographystyle{IEEEtran}

\end{document}